# Decoding AI Authorship: Can LLMs Truly Mimic Human Style Across Literature and Politics?

Nasser A Alsadhan

College of Computer and Information Sciences, King Saud University, Riyadh, Saudi Arabia
nsadhan@ksu.edu.sa

# Abstract

Amidst the rising capabilities of generative AI to mimic specific human styles, this study investigates the ability of state-of-the-art large language models (LLMs), including GPT-4o, Gemini 1.5 Pro, and Claude Sonnet 3.5, to emulate the authorial signatures of prominent literary and political figures: Walt Whitman, William Wordsworth, Donald Trump, and Barack Obama. Utilizing a zero-shot prompting framework with strict thematic alignment, we generated synthetic corpora evaluated through a complementary framework combining transformer-based classification (BERT) and interpretable machine learning (XGBoost). Our methodology integrates Linguistic Inquiry and Word Count (LIWC) markers, perplexity, and readability indices to assess the divergence between AI-generated and human-authored text. Results demonstrate that AI-generated mimicry remains highly detectable, with XGBoost models trained on a restricted set of eight stylometric features achieving accuracy comparable to high-dimensional neural classifiers. Feature importance analyses identify perplexity as the primary discriminative metric, revealing a significant divergence in the stochastic regularity of AI outputs compared to the higher variability of human writing. While LLMs exhibit distributional convergence with human authors on low-dimensional heuristic features, such as syntactic complexity and readability, they do not yet fully replicate the nuanced affective density and stylistic variance inherent in the human-authored corpus. By isolating the specific statistical gaps in current generative mimicry, this study provides a comprehensive benchmark for LLM stylistic behavior and offers critical insights for authorship attribution in the digital humanities and social media.

# 1. Introduction

Generative models powered by large language models (LLMs) have significantly impacted fields ranging from academic publishing to creative writing (Castillo 2025; Wu et al. 2025). Their ability to produce text with human-like fluency has opened new opportunities for co-creation while simultaneously raising concerns regarding authorship, originality, and the potential for misuse in digital environments (Wu et al. 2025). For instance, the risk of models reproducing copyrighted material or mimicking the distinct 'voice' of an author without consent has introduced complex ethical and legal questions regarding intellectual property and authenticity (Lucchi 2024; Iglesias et al. 2025). Beyond legal frameworks, AI-generated text also carries risks of deception, such as impersonating individuals or spreading misinformation, underscoring the need for critical evaluation of these technologies in creative domains (Mitchell et al. 2023).

From a technical perspective, state-of-the-art models such as GPT-4 and Gemini Pro demonstrate high proficiency in generating grammatically coherent and contextually relevant content. However, recent studies in computational stylometry suggest that the capacity of LLMs to fully emulate the implicit and nuanced stylistic qualities of specific individuals remains limited (Wang et al. 2025; O'Sullivan 2025). Elements deeply tied to authorial identity, such as idiosyncratic rhythm, syntactic variety, and thematic depth, pose significant challenges for current generative architectures, which often default to a more standardized or 'average' linguistic profile (Sadasivan et al. 2023; Wu et al. 2025). Understanding these potential limitations is crucial, as the degree of stylistic emulation has significant implications not only for Natural Language Processing (NLP) but also for literary studies, education, and the security of digital discourse.

The use of LLMs for stylistic imitation presents a dualistic challenge. While these systems can democratize access to complex literary forms, serving as pedagogical tools and aids for cultural preservation, the ability to convincingly mimic recognizable voices poses risks of misattribution and the potential erosion of traditional authorial boundaries (Fenwick et al 2023). Such issues are amplified in an era where distinguishing between authentic human voices and synthetic imitations, often described as the 'Turing Test for Style,' is increasingly difficult for human readers alone (Mikros 2025, Uchendu, Le, and Lee 2023).

Quantitative analysis suggests that while current LLMs achieve high levels of fluency, they appear to struggle to replicate the heterogeneous diversity of human expression. Stylometric measures like Burrows' Delta reveal that LLM outputs exhibit a high degree of stylistic uniformity, resulting in tight clusters that remain statistically distinct from the broader, more varied stylistic fingerprints characteristic of human-authored creative writing (O'Sullivan 2025). This gap suggests that machine-generated text can often be identified through the unique structure of a model's probability function, specifically, the tendency for LLM-sampled text to occupy regions of negative curvature that human writing does not (Mitchell et al. 2023). Such stochastic regularities suggest that information-theoretic metrics, such as perplexity, may serve as primary indicators of synthetic origin. Investigating these limitations is not only a matter of technical benchmarking but also essential for understanding the current state of AI-driven creativity. However, existing studies primarily examine aggregate stylistic separability or detection performance, offering comparatively limited insight into how psycholinguistic, readability, and stylistic features behave under systematically controlled, author-specific prompting across multiple models and genres.

This study investigates the extent of stylistic mimicry by examining LLM-generated imitations of four prominent figures: Walt Whitman, William Wordsworth, Donald Trump, and Barack Obama. These figures were selected to represent contrasting poles of stylistic expression, ranging from the high lexical density and metaphorical complexity of 19th-century poetic verse to the structural sparsity and imperative nature of modern political rhetoric. This selection provides a diverse testbed for authorship detection across varying linguistic constraints. The analysis integrates quantitative metrics, including perplexity scores, readability indices, and psycholinguistic features extracted via Linguistic Inquiry and Word Count (LIWC), to provide a multi-dimensional assessment of AI mimicry.

The contribution of this study is threefold. First, it evaluates the efficacy of transformer-based and feature-driven classifiers in distinguishing human from AI authorship across different genres. Second, it provides a fine-grained analysis of how stylistic and psycholinguistic patterns diverge between individual human authors and their AI counterparts. Third, it introduces an interpretable XGBoost architecture that identifies the specific linguistic markers most indicative of AI authorship. By demonstrating that a compact, 8-dimensional feature set can achieve performance comparable to high-dimensional neural baselines, this study contributes to the broader goal of responsible and transparent AI deployment.

## 2. Literature Review

In recent years LLMs have demonstrated remarkable capabilities in natural language generation, challenging traditional assumptions about human-exclusive cognitive and creative skills. These models, such as GPT (Gallifant 2024), and Gemini (Team G et al. 2024), have increasingly been investigated for their potential to emulate aspects of human cognition, including decision-making, problem-solving, and language understanding. Zhuang et al. (2023) examined the extent to which LLMs can approximate human performance in cognition-related tasks, such as the Wason Selection Task and Raven-like matrices. Their findings suggest that while LLMs can achieve near-human competency in structured cognitive tasks, their performance is constrained by the models' design frameworks and pretraining data. This proficiency in structured tasks suggests a potential, yet untested, capacity for emulating more nuanced and subjective aspects of human authorship, such as stylistic signatures.

Hagendorff (2023) investigated LLMs' engagement with System 1 and System 2 thinking, distinguishing between fast intuitive processes and slower deliberate reasoning. Their study indicates that while models can utilize contextual information to perform reasoning tasks, their output is heavily dependent on model architecture rather than genuine understanding. This distinction is crucial in the context of stylistic imitation, as it implies that while LLMs may replicate surface-level patterns of an author's writing, deeper interpretive choices, including thematic coherence and rhetorical strategies, appear more constrained. Consistent with this, recent stylometric studies suggest that LLM-generated texts tend to exhibit reduced idiosyncratic rhythm and lower stylistic diversity when compared to human-authored creative writing (O'Sullivan 2025; Sadasivan et al. 2023).

Creativity, a vital component of authorship, has also been examined. Stevenson et al. (2022) used the Remote Associates Test to evaluate GPT-3's creative potential, finding that it could generate plausible solutions and demonstrate functional creativity. However, when considering the emulation of individual authorial style, capturing idiosyncratic preoccupations goes beyond surface-level grammar. This has led to the emergence of what Mikros (2025) and Uchendu, Le, and Lee (2023) describe as the 'Turing Test for Style,' where the challenge shifts from generating coherent text to maintaining a believable authorial 'fingerprint' that can withstand quantitative scrutiny.

While LLMs can recognize emotional states in controlled settings (Schueller and Morris, 2023), their performance in open-ended, author-specific mimicry remains comparatively less examined in the literature. Current detection approaches primarily emphasize binary accuracy, distinguishing 'human' from 'machine,' often by introducing benchmark datasets and evaluating a range of classifiers (Al Bataineh et al. 2025). These methods typically utilize surface-level linguistic signals like lexical distributions (Wang et al. 2025; Wu et al. 2025). Related stylometric analyses further indicate that AI-generated texts exhibit systematic differences in readability, lexical density, and syntactic structure across models (Jaashan and Bin-Hady 2025). However, as Mitchell et al. (2023) demonstrate through 'DetectGPT,' these models tend to occupy distinctive regions of probability curvature. This suggests that the information-theoretic 'signature' of a model, often manifested as localized perplexity spikes or regularities, remains a more reliable indicator than lexical choice alone. Additional work has raised concerns regarding the robustness of these detection systems, particularly when texts are stylistically modified or paraphrased (Makinde et al. 2025).

Taken together, these insights, an underexplored gap emerges: many prior studies emphasize aggregate stylistic tendencies or detection performance (Fariello et al. 2025), offering comparatively less insight into the stability of stylistic features when models are subjected to adversarial, zero-shot prompting designed to mimic specific individuals across genres and model architectures (O'Sullivan, 2025). The present study adopts a complementary perspective: rather than optimizing detection accuracy, it systematically examines how stylistic, psycholinguistic, and readability features manifest in AI-generated texts conditioned on specific literary and political authors.

Several techniques have been developed for quantifying stylistic and psycholinguistic features in text. The LIWC tool and a range of readability metrics enable detailed analysis of linguistic, cognitive, and social patterns in writing (Pennebaker, Francis, and Booth 2001; Flesch, 1948;

Gunning, 1952; Kincaid et al., 1975). Together, these methods provide a robust framework for assessing surface-level stylistic variation in both human-authored and AI-generated text. Building on this framework, the present study integrates these tools with an interpretable XGBoost model to move beyond black-box detection, identifying linguistic markers, such as perplexity and psycholinguistic shifts, which distinguish consistent stylistic tendencies of human authors from their synthetic imitations.

# 3. Methodology

The methodology of this study is designed to systematically examine the ability of LLMs to mimic distinctive authorial styles beyond surface-level fluency. The approach, illustrated in Figure 1, integrates multiple stages, including dataset curation, model selection and prompt engineering, corpus preprocessing, predictive modeling, and combined linguistic and psycholinguistic analysis. By moving from 'black-box' classification to feature-based interpretability, this framework addresses the need for a more nuanced understanding of AI-generated stylistic fingerprints (Sadasivan et al. 2023).

## 3.1 Dataset Collection and Preprocessing

To create a robust experimental foundation, we curated datasets from two literary figures, Walt Whitman and William Wordsworth (PeterS111 2022), and two political figures, Donald Trump and Barack Obama (codebreaker619 2020; Gajare 2021). These individuals were selected to represent contrasting stylistic regimes: highly aesthetic, expressive literary verse (Whitman, Wordsworth) versus constrained, institutional political discourse (Trump, Obama), enabling analysis across distinct stylistic and communicative domains. The literary corpus was compiled using established open-source poetry repositories to ensure high-quality, representative samples of each author's work.

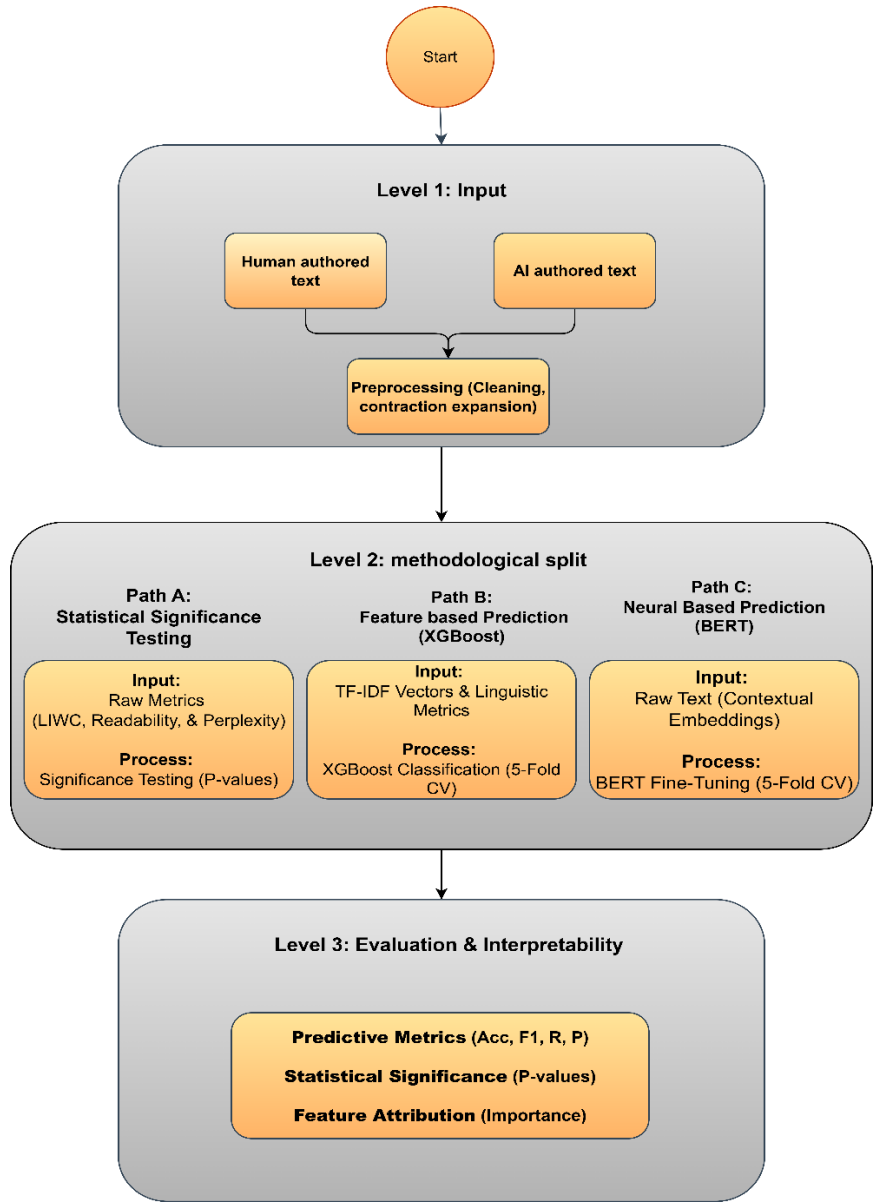

**Figure 1:** Pipeline for assessing LLMs stylistic fidelity.

Tokenization using the NLTK library was applied for corpus-level linguistic and psycholinguistic analysis (e.g. LIWC and readability metrics), as well as for generating the term frequency–inverse document frequency (TF-IDF) vectorization utilized by the XGBoost classifier. Importantly, raw text strings were provided to both the BERT classifier and the GPT-2 Large model for perplexity computation, allowing each to apply its native tokenization mechanisms. This approach was essential to avoid interference with learned representations and to preserve the mathematical integrity of the models' original probability distributions. No lemmatization or stop-word removal was applied to the corpus; this was a deliberate choice to preserve the idiosyncratic syntactic markers, function-word distributions, and rhythmic patterns essential for stylometric analysis (O'Sullivan 2025). Standardization included converting contractions (e.g. 'don't' → 'do not') to maintain formal consistency. Texts were sampled from each author's available corpus without temporal filtering to preserve natural stylistic variability. Because stylistic and lexical diversity metrics are sensitive to text length, sample-level and author-level counts were monitored for genre alignment. Specifically, the literary corpus was restricted to 300 poems per author, while the political corpus was maintained at 3,000 tweets per author. Descriptive statistics, including total token counts, mean word length, mean tokens per text, and Type-Token Ratio (TTR), are reported in Table 1 to contextualize subsequent analyses.

**Table 1: Original Dataset Summary**

| Author | Text Type | Quantity | Total Tokens | Mean Word Length | Mean Tokens / Text | Type-Token Ratio |
|---|---|---|---|---|---|---|
| Walt Whitman | Poems | 300 | 64,067 | 4.7 | 213 | 0.115 |
| William Wordsworth | Poems | 300 | 60,098 | 4.5 | 200 | 0.110 |
| Donald Trump | Tweets | 3,000 | 56,270 | 4.6 | 18 | 0.090 |
| Barack Obama | Tweets | 3,000 | 45,528 | 5.3 | 15 | 0.092 |

**3.2 Large Language Model Selection and Prompt Design**

We employed three state-of-the-art Large Language Models (LLMs), GPT-4o (Gallifant 2024), Claude 3.5 Sonnet (Anthropic 2025), and Gemini 1.5 Pro (Team G et al. 2024), to produce synthetic texts mimicking the target authors. These models were selected to represent distinct contemporary design paradigms, specifically variations in Reinforcement Learning from Human Feedback (RLHF) and architectural scaling, ensuring that our findings reflect broader trends in generative AI rather than model-specific artifacts. Following the prompt engineering challenges identified by Mitchell et al. (2023), we utilized a minimalist zero-shot prompting framework intentionally designed to isolate each model's intrinsic stylistic behavior and latent representations, thereby mitigating the risk of researcher-induced bias associated with complex instruction-following or few-shot priming. Furthermore, while we acknowledge that proprietary system prompts vary across providers, our study evaluates these models as integrated architectural units, capturing the total operational style of the model as delivered to end-users.

While we initially considered prompts incorporating specific stylistic parameters (e.g., instructions regarding archaic diction or sentence structure), we ultimately standardized a minimalist template across all authors and models to prevent researcher-induced stylistic bias and ensure that observed patterns emerged from the models' internal training data. These templates consisted of: 'Write a poem exclusively in the style of [Author Name] on the topic of [Thematic Keyword]' for the literary corpus and 'Write a tweet exclusively in the style of [Author Name] about [Policy Topic]' for the political corpus.

All generations were conducted between December 2024 and March 2025 via official API endpoints using a Temperature setting of 1.0 to preserve the models' original probability distributions without artificial sharpening. This choice ensures that our stylistic analysis captures the models' baseline stochasticity and prevents the artificial inflation of classification accuracy associated with lower-entropy, deterministic output states. To ensure that the classifiers distinguished style rather than content, we enforced strict thematic alignment with the human-authored ground truth. For the literary texts, 300 poems per author were generated using keywords derived from the primary source metadata; for the political texts, 3,000 tweets per author were restricted to dominant shared policy domains, specifically economics and healthcare. This established a consistent semantic baseline, preventing classification models from leveraging subject-specific lexical cues and forcing a reliance on idiosyncratic stylistic markers.

To ensure that classification models learned stylistic rather than conversational cues, automatically generated meta-responses (e.g., prefatory phrases such as 'Sure, here is…') were removed using a lightweight post-processing filter. No lexical, syntactic, or semantic content was otherwise altered. The resulting synthetic datasets are summarized in Table 2.

Table 2: Synthetic Dataset Summary

| Author | Text Type | LLMs Used | Total Tokens | Mean Word Length | Type-Token Ratio | Mean Tokens / Text | Quantity per LLM | Total Quantity |
|---|---|---|---|---|---|---|---|---|
| Walt Whitman | Poems | GPT-4o | 68,131 | 5.0 | 0.102 | 227 | 300 | 900 |
| | | Gemini 1.5 Pro | 76,394 | 4.9 | 0.082 | 254 | | |
| | | Sonnet 3.5 | 61,164 | 5.2 | 0.054 | 203 | | |
| William Wordsworth | Poems | GPT-4o | 60,948 | 4.8 | 0.090 | 203 | 300 | 900 |
| | | Gemini 1.5 Pro | 65,209 | 4.9 | 0.078 | 217 | | |
| | | Sonnet 3.5 | 57,326 | 5.3 | 0.052 | 191 | | |
| Donald Trump | Tweets | GPT-4o | 54,302 | 5.3 | 0.025 | 18 | 3,000 | 9,000 |
| | | Gemini 1.5 Pro | 57,433 | 5.6 | 0.011 | 19 | | |

| | | Sonnet 3.5 | 49,511 | 5.3 | 0.028 | 16.5 | | |
| **Barack Obama** | Tweets | GPT-4o | 50,203 | 5.5 | 0.013 | 16 | 3,000 | 9,000 |
| | | Gemini 1.5 Pro | 49,101 | 5.6 | 0.014 | 16.3 | | |
| | | Sonnet 3.5 | 45,821 | 5.2 | 0.016 | 15.3 | | |

## 3.3 Linguistic and Psycholinguistic Analysis

To evaluate stylistic and psycholinguistic properties of both human-authored and AI-generated texts, a range of linguistic metrics was computed. Readability and fluency were assessed using Flesch Reading Ease, Flesch–Kincaid Grade Level, and the Gunning Fog Index, which capture complementary dimensions of textual complexity and are widely used in stylometric and educational research. Readability measures primarily designed for longer passages were not emphasized, as the corpus includes short-form texts for which such metrics are less stable. Several indices were found to be highly correlated; therefore, a representative subset was retained to reduce redundancy while preserving interpretability.

Perplexity (PPL) was computed using the pretrained GPT-2 Large language model (774M parameters) via the HuggingFace Transformers library. GPT-2 Large provides a stable, general-purpose probability distribution over modern English due to its training on the OpenWebText corpus and remains a standard benchmark for perplexity estimation. Crucially, by utilizing a pre-trained model independent of the generative systems used in this study (GPT-4o, Sonnet 3.5, and Gemini 1.5 Pro), we ensure that the perplexity scores measure the intrinsic statistical complexity of the texts rather than artifacts of data overlap between the generator and the evaluator.

Psycholinguistic features were extracted using LIWC-22. All LIWC categories were computed; however, analytical emphasis was placed on dimensions that exhibited consistent and interpretable variation across authors and models. In particular, tone, authenticity, analytic thinking, and social presence were examined in detail, as these dimensions are closely tied to authorial stance, emotional expression, and interpersonal orientation. Reporting all LIWC dimensions in full would substantially increase result complexity without proportionate interpretive gain.

Several linguistic features are known to be partially correlated, reflecting overlapping definitions of textual complexity. This correlation was expected and does not undermine the present analysis. Descriptive comparisons do not rely on coefficient-based inference, and the feature-based predictive modeling employs a tree-based classifier that is robust to correlated predictors. Accordingly, feature importance is interpreted comparatively rather than causally. Full details of the extracted features are provided in Table 3.

Statistical significance for all linguistic and psycholinguistic comparisons was assessed using the Mann–Whitney U test. This non-parametric test was selected to account for the distributional characteristics of stylistic and psycholinguistic features, which may deviate from normality and exhibit substantial variability across authors, genres, and generation conditions. In line with non-parametric reporting standards, results are summarized using the median and interquartile range (IQR), which provide a robust characterization of central tendency and dispersion for skewed or heterogeneous distributions. Statistical significance was evaluated using a threshold of $\alpha = .05$; values reaching the lower limit of numerical precision are reported as $p < .001$

Table 3: Evaluation Metrics, Formulas, and Interpretation

| Metric | What It Measures | High Score Interpretation | Low Score Interpretation |
| --- | --- | --- | --- |
| Tone (Pennebaker, 2001) | Emotional valence (positive vs. negative tone) | Cheerful, positive emotional expression | Sad, anxious, angry, or negative tone |
| Authentic (Pennebaker, 2001) | Sincerity, honesty, and personal disclosure | Genuine, self-reflective, personal writing | Formal, distant, or impersonal tone |
| Clout (Pennebaker, 2001) | Social status, confidence, or leadership conveyed | Confident, authoritative, high-status voice | Hesitant, insecure, deferential tone |
| Analytic (Pennebaker, 2001) | Logical, hierarchical, and structured thinking | Formal, academic, analytical writing | Intuitive, narrative, or conversational writing |
| Flesch Reading Ease (Flesch, 1948) | How easy a text is to read (0–100 scale). Calculated from number of words, sentences, and syllables | Very easy to read (simple words, short sentences) | Very hard to read (complex sentences and vocabulary) |
| Flesch-Kincaid Grade Level (Flesch, 1975) | Approximate U.S. school grade level required to understand the text. Based on sentence length and word complexity | High grade level indicates complex, academic writing | Low grade level indicates simple, accessible writing |
| Gunning Fog Index (Gunning, 1952) | Years of formal education needed to understand the text, based on sentence length and proportion of complex words | Complex writing requiring more education (>12 = college-level) | Simple, clear writing requires less education (<8 = general audience) |
| Perplexity | Predictive uncertainty of a language model on the text; measures how well the model predicts the next word | High perplexity indicates unexpected, irregular, or less fluent text | Low perplexity indicates predictable, fluent text |

## 3.4 Predictive Modeling

To evaluate stylistic distinguishability between human-authored and AI-generated texts, we employed a complementary predictive modeling design across three experimental conditions. This approach compares raw lexical choice, deep neural representations, and interpretable psycholinguistic features. To ensure the highest level of methodological rigor and to prevent bias from idiosyncratic data partitioning, we implemented a class-balanced cross-validation protocol. This architecture is particularly critical for maintaining the validity of results across our varying dataset sizes, which range from 600 samples for the poetic datasets to 6,000 for the political datasets.

For the transformer-based approach, we fine-tuned the bert-base-uncased model utilizing raw text sequences as the primary input for binary authorship classification. Training followed established benchmarks (Devlin et al. 2019) using the AdamW optimizer with a learning rate of 2e-5 and a batch size of 32 over three epochs. To ensure the results represent the model's generalizable discriminative power rather than the results of idiosyncratic tuning, we utilized a stratified 5-fold cross-validation scheme with these fixed hyperparameters. In each fold, 20% of the data was held out as a strictly unseen test set. Of the remaining 80% used for training, a 10% internal validation subset was reserved to monitor convergence and prevent overfitting. This conservative approach, prioritizing standard configurations over extensive hyperparameter optimization, minimizes the risk of over-fitting to the specific lexical nuances of the training folds while providing a robust estimate of the authorship signal.

In parallel, we utilized the XGBoost algorithm (Chen and Guestrin 2016) to isolate the source of the authorship signal through two distinct feature sets. The lexical experiment utilized TF-IDF vectorized word counts to establish a baseline for vocabulary-based discrimination, while the stylometric experiment utilized only normalized numeric features represented as discrete numerical vectors, including readability indices, perplexity scores, and LIWC-based psycholinguistic measures. To maintain stability during the tuning phase, the XGBoost models were evaluated using a stratified nested cross-validation protocol. An inner 5-fold cross-validation loop was used for hyperparameter optimization via grid search, specifically targeting maximum tree depth (3, 6) and learning rate (0.01, 0.1), while an independent outer 5-fold cross-validation loop was used to report the final performance metrics. By separating the tuning phase from the evaluation phase, this architecture provides a strictly conservative and statistically robust measure of authorship signal. We report the mean and standard deviation of accuracy, precision, recall, and F1-scores across the outer folds as the definitive estimate of discriminative power across all experimental conditions.

# 4 Results and Discussion

This section reports the findings of our investigation into the extent to which large language models can reproduce stable, author-specific stylistic patterns beyond surface-level fluency. We first analyze linguistic and psycholinguistic features, including LIWC dimensions, readability indices, and perplexity, to examine how AI-generated texts compare to human-authored baselines in terms of stylistic and distributional properties. We then report the performance of predictive models across three complementary settings: (i) BERT classifiers trained on raw contextual embeddings, (ii) XGBoost classifiers trained on interpretable stylistic and psycholinguistic features, and (iii) XGBoost classifiers trained on TF–IDF representations to establish a vocabulary-based baseline. Together, these analyses evaluate stylistic convergence and divergence across poetic and political domains. Following the methodology detailed in Section 3, all human vs. AI comparisons were evaluated using the Mann–Whitney U test, with results reported via the median and interquartile range (IQR).

### 4.1 Linguistic and Psycholinguistic Metrics

The analysis initially examines LIWC dimensions (Tables 4–7), which provide a window into the psychological and structural orientation of the texts, followed by an assessment of perplexity and readability indices (Tables 8–11). The results of the LIWC analysis for Walt Whitman and the corresponding LLM-generated texts are shown in Table 4. These scores span the categories of Analytic, Clout, Authentic, and Tone, and serve as a basis for comparing stylistic and psychological patterns between human and AI writing.

**Table 4:** Linguistic Inquiry and Word Count Results (Whitman)

| Metric | Human Median (IQR) | GPT-4O Median (IQR) | GPT-4O p-value | Gemini 1.5 Pro Median (IQR) | Gemini 1.5 Pro p-value | Sonnet 3.5 Median (IQR) | Sonnet 3.5 p-value |
|---|---|---|---|---|---|---|---|
| Analytic | 80.13 (32.70) | 89.32 (24.00) | < .001 | 88.35 (25.11) | < .001 | 85.85 (13.96) | < .001 |
| Clout | 49.06 (47.88) | 45.22 (49.22) | < .05 | 44.06 (49.25) | n.s. | 73.40 (53.93) | < .001 |
| Authentic | 60.48 (47.30) | 88.42 (21.43) | < .001 | 84.35 (26.63) | < .001 | 92.64 (13.28) | < .001 |
| Tone | 51.47 (52.47) | 65.81 (32.27) | < .001 | 45.29 (33.73) | < .05 | 71.91 (27.29) | < .001 |

Whitman's originals scored a median of 80.13 on the Analytic dimension. In comparison, GPT-4o (89.32), Gemini 1.5 Pro (88.35), and Sonnet 3.5 (85.85) produced significantly higher medians (p < .001). For Clout, Whitman's median was 49.06. Gemini 1.5 Pro (44.06) showed no statistically significant difference from this baseline (p > 0.05), while GPT-4o (45.22, p < .05) and Sonnet 3.5 (73.40, p < .001) both deviated significantly. In terms of Authenticity, Whitman's median was 60.48, while all LLM samples exhibited significantly higher values (p < .001): GPT-4o (88.42), Gemini 1.5 Pro (84.35), and Sonnet 3.5 (92.64). Finally, Whitman's original Tone median (51.47) differed significantly from all tested models (p < .05 or p < .001), with GPT-4o and Sonnet 3.5 scoring higher and Gemini 1.5 Pro scoring lower (45.29).

**Table 5:** Linguistic Inquiry and Word Count Results (Wordsworth)

| Metric | Human Median (IQR) | GPT-4O Median (IQR) | GPT-4O p-value | Gemini 1.5 Pro Median (IQR) | Gemini 1.5 Pro p-value | Sonnet 3.5 Median (IQR) | Sonnet 3.5 p-value |
|---|---|---|---|---|---|---|---|
| Analytic | 79.10 (23.62) | 89.52 (15.11) | < .001 | 83.27 (23.45) | < .05 | 86.19 (10.84) | < .001 |
| Clout | 61.53 (32.87) | 61.21 (43.00) | n.s. | 46.90 (47.07) | < .001 | 63.25 (37.22) | n.s. |
| Authentic | 25.70 (35.05) | 80.70 (26.45) | < .001 | 77.65 (32.91) | < .001 | 91.47 (14.50) | < .001 |
| Tone | 62.75 (55.80) | 85.02 (28.95) | < .001 | 61.14 (51.05) | n.s. | 85.43 (17.44) | < .001 |

Table 5 presents the LIWC analysis for William Wordsworth. Wordsworth's poems scored a median of 79.10 on Analytic thinking. In contrast, the LLMs produced significantly higher medians: GPT-4o (89.52), Sonnet 3.5 (86.19), and Gemini 1.5 Pro (83.27), with all results reaching statistical significance (p < .05). For Clout, the human median was 61.53. GPT-4o (61.21) and Sonnet 3.5 (63.25) showed no statistically significant difference from this baseline (p > 0.05), while Gemini 1.5 Pro produced a significantly lower median of 46.90 (p < .001). The Authentic dimension showed a significant difference across all samples: Wordsworth's median was 25.70, while GPT-4o (80.70), Gemini 1.5 Pro (77.65), and Sonnet 3.5 (91.47) all exhibited significantly higher values (p < .001). Finally, Tone medians were significantly higher in the GPT-4o (85.02) and Sonnet 3.5 (85.43) samples compared to the human median of 62.75 (p < .001), though the Gemini 1.5 Pro median (61.14) was not statistically different from the original (p > 0.05).

**Table 6:** Linguistic Inquiry and Word Count Results (Trump)

| Metric | Human Median (IQR) | GPT-4O Median (IQR) | GPT-4O p-value | Gemini 1.5 Pro Median (IQR) | Gemini 1.5 Pro p-value | Sonnet 3.5 Median (IQR) | Sonnet 3.5 p-value |
|---|---|---|---|---|---|---|---|
| Analytic | 63.53 (59.15) | 45.12 (44.02) | < .001 | 39.70 (36.59) | < .001 | 67.11 (28.31) | n.s. |
| Clout | 62.84 (74.59) | 84.23 (37.14) | < .001 | 60.57 (58.27) | < .01 | 50.89 (39.68) | < .001 |
| Authentic | 17.84 (66.84) | 26.78 (63.34) | < .001 | 10.18 (33.99) | < .001 | 16.00 (34.88) | n.s. |
| Tone | 20.23 (78.77) | 75.77 (78.42) | < .001 | 3.20 (19.23) | < .001 | 33.54 (44.97) | < .001 |

Table 6 compares LIWC scores for Donald Trump's original tweets with those generated by GPT-4o, Gemini 1.5 Pro, and Sonnet 3.5. Trump's original tweets show a median Analytic score of 63.53. GPT-4o (45.12) and Gemini 1.5 Pro (39.70) significantly reduced this structural complexity (p < .001), while Sonnet 3.5 (67.11, p > 0.05) showed no statistically significant difference from the human baseline.

For Clout, Trump's baseline of 62.84 was sharply inflated by GPT-4o (84.23, p < .001), suggesting an over-exaggeration of his authoritative stance. Conversely, Sonnet 3.5 (50.89, p < .001) significantly understated his typical level of confidence, while Gemini 1.5 Pro (60.57, p < .01) remained the most proximal to the original median. Trump's Authenticity median was 17.84; while GPT-4o (26.78) and Gemini 1.5 Pro (10.18) showed significant deviations (p < .001), Sonnet 3.5 (16.00, p > 0.05) successfully replicated this specific dimension. Finally, Trump's Tone (20.23) was vastly over-estimated by GPT-4o (75.77, p < .001) and starkly under-estimated by Gemini 1.5 Pro (3.20, p < .001).

**Table 7:** Linguistic Inquiry and Word Count Results (Obama)

| Metric | Human Median (IQR) | GPT-4O Median (IQR) | GPT-4O p-value | Gemini 1.5 Pro Median (IQR) | Gemini 1.5 Pro p-value | Sonnet 3.5 Median (IQR) | Sonnet 3.5 p-value |
|---|---|---|---|---|---|---|---|
| Analytic | 80.96 (49.81) | 66.04 (42.02) | < .001 | 62.10 (36.15) | < .001 | 80.20 (22.41) | < .05 |
| Clout | 79.51 (58.69) | 94.84 (26.13) | < .001 | 92.24 (17.43) | < .001 | 63.73 (45.19) | < .001 |
| Authentic | 19.26 (68.37) | 8.42 (38.59) | < .001 | 3.06 (23.32) | < .001 | 7.47 (18.40) | < .001 |
| Tone | 20.23 (76.51) | 96.74 (26.55) | < .001 | 88.66 (28.09) | < .001 | 78.39 (44.39) | < .001 |

Table 7 presents the LIWC results for Barack Obama's original tweets alongside the synthetic texts generated by GPT-4o, Gemini 1.5 Pro, and Sonnet 3.5. Obama's tweets scored a median of 80.96 on Analytic thinking. While Sonnet 3.5 (80.20) was the most proximal to this baseline, the difference remained statistically significant (p < .05). GPT-4o (66.04) and Gemini 1.5 Pro (62.10) exhibited even greater significant reductions in this dimension (p < .001).

For Clout, the original human baseline of 79.51 was significantly exceeded by GPT-4o (94.84) and Gemini 1.5 Pro (92.24), while Sonnet 3.5 (63.73) scored significantly lower than the human median (p < .001 for all). The Authentic dimension showed a universal significant decrease across all models: Obama's human median of 19.26 was significantly higher than the medians produced by GPT-4o (8.42), Sonnet 3.5 (7.47), and Gemini 1.5 Pro (3.06) (p < .001 for all). Finally, the Tone dimension revealed significant distributional shifts toward higher positivity in the generated texts. While Obama's original tweets registered a median of 20.23, all models produced significantly higher medians (p < .001), with GPT-4o reaching 96.74.

**Table 8:** Perplexity and Readability for Whitman

| Metric | Human Median (IQR) | GPT-4O Median (IQR) | GPT-4O p-value | Gemini 1.5 Pro Median (IQR) | Gemini 1.5 Pro p-value | Sonnet 3.5 Median (IQR) | Sonnet 3.5 p-value |
|---|---|---|---|---|---|---|---|
| perplexity | 82.10 (48.83) | 58.74 (28.85) | < .001 | 69.16 (34.78) | < .001 | 83.47 (22.63) | n.s. |
| Flesch reading ease | 52.60 (55.25) | 47.29 (40.50) | n.s. | 61.84 (27.92) | < .001 | 57.95 (15.07) | < .001 |

| | | | | | | |
|---|---|---|---|---|---|---|
| Flesch Kincaid grade | 13.70 (18.95) | 15.60 (16.05) | n.s. | 11.80 (9.30) | < .001 | 11.50 (5.03) | < .001 |
| Gunning fog | 16.13 (20.11) | 17.74 (16.20) | n.s. | 13.71 (9.25) | < .001 | 13.36 (5.01) | < .001 |

Table 8 presents the perplexity and readability metrics for original poems by Walt Whitman compared to the LLM-generated texts. Whitman's originals exhibited a median perplexity of 82.10. GPT-4o (58.74) and Gemini 1.5 Pro (69.16) yielded significantly lower perplexity scores (p < .001), while Sonnet 3.5 (83.47) showed no statistically significant difference from the human baseline (p > 0.05). Regarding readability, Whitman's poems had a median Flesch Reading Ease score of 52.60 and a Flesch-Kincaid Grade Level of 13.70. GPT-4o did not differ significantly in these metrics (p > 0.05). However, Gemini 1.5 Pro and Sonnet 3.5 produced significantly higher Reading Ease scores (lower complexity) and lower Grade Levels (p < .001). The Gunning Fog Index for Whitman (16.13) followed a similar pattern, with Gemini 1.5 Pro (13.71) and Sonnet 3.5 (13.36) showing significantly lower values (p < .001), while GPT-4o (17.74) remained statistically comparable (p > 0.05).

**Table 9:** Perplexity and Readability for Wordsworth

| Metric | Human Median (IQR) | GPT-4O Median (IQR) | GPT-4O p-value | Gemini 1.5 Pro Median (IQR) | Gemini 1.5 Pro p-value | Sonnet 3.5 Median (IQR) | Sonnet 3.5 p-value |
|---|---|---|---|---|---|---|---|
| perplexity | 110.80 (52.74) | 61.91 (31.55) | < .001 | 99.95 (47.12) | < .001 | 111.29 (33.41) | n.s. |
| Flesch reading ease | 68.13 (18.01) | 72.04 (12.39) | < .001 | 68.28 (20.11) | n.s. | 54.15 (11.98) | < .001 |
| Flesch Kincaid grade | 10.80 (5.80) | 9.70 (3.90) | < .001 | 10.90 (5.05) | n.s. | 12.70 (3.43) | < .001 |
| Gunning fog | 12.93 (5.81) | 11.73 (3.78) | < .001 | 12.99 (5.34) | n.s. | 14.54 (3.68) | < .001 |

Table 9 presents the metrics for William Wordsworth. The original poems exhibit a median perplexity of 110.80. GPT-4o (61.91) and Gemini 1.5 Pro (99.95) exhibited significantly lower perplexity (p < .001), whereas Sonnet 3.5 (111.29) showed no significant difference from the original human texts (p > 0.05). Wordsworth's originals scored 68.13 on Flesch Reading Ease and 10.80 on Grade Level. GPT-4o produced significantly higher readability (lower complexity) scores (p < .001), while Sonnet 3.5 yielded significantly lower readability (higher complexity) scores (p < .001). Gemini 1.5 Pro remained statistically aligned with the original human baseline in all readability dimensions (p > 0.05).

**Table 10:** Perplexity and Readability for Trump

| Metric | Human Median (IQR) | GPT-4O Median (IQR) | GPT-4O p-value | Gemini 1.5 Pro Median (IQR) | Gemini 1.5 Pro p-value | Sonnet 3.5 Median (IQR) | Sonnet 3.5 p-value |
|---|---|---|---|---|---|---|---|
| perplexity | 65.27 (59.51) | 38.32 (21.98) | < .001 | 52.18 (22.68) | < .001 | 31.92 (13.35) | < .001 |
| Flesch reading ease | 66.74 (36.89) | 65.73 (18.24) | n.s. | 64.57 (18.74) | n.s. | 64.41 (11.84) | < .01 |
| Flesch kincaid grade | 7.40 (6.10) | 6.30 (2.70) | < .001 | 6.30 (3.20) | < .001 | 8.00 (2.70) | < .001 |
| Gunning fog | 8.28 (5.71) | 8.01 (3.08) | < .001 | 7.49 (1.85) | < .001 | 9.88 (2.77) | < .001 |

Table 10 presents the metrics for Donald Trump. The original tweets exhibit a median perplexity of 65.27. All three LLMs, GPT-4o (38.32), Gemini 1.5 Pro (52.18), and Sonnet 3.5 (31.92), produced significantly lower perplexity scores (p < .001). On the Flesch Reading Ease scale, Trump's median was 66.74. GPT-4o and Gemini 1.5 Pro showed no significant difference (p > 0.05), while Sonnet 3.5 yielded a significantly lower score (64.41, p < .01). Trump's Grade Level median of 7.40 differed significantly from all models (p < .001), with GPT-4o and Gemini 1.5 Pro scoring lower (6.30) and Sonnet 3.5 scoring higher (8.00). The Gunning Fog Index followed this trend, with Sonnet 3.5 (9.88) being significantly higher than the human baseline (8.28, p < .001).

**Table 11:** Perplexity and Readability for Obama

| Metric | Human Median (IQR) | GPT-4O Median (IQR) | GPT-4O p-value | Gemini 1.5 Pro Median (IQR) | Gemini 1.5 Pro p-value | Sonnet 3.5 Median (IQR) | Sonnet 3.5 p-value |
|---|---|---|---|---|---|---|---|
| perplexity | 63.83 (55.97) | 27.60 (19.58) | < .001 | 25.42 (16.04) | < .001 | 17.92 (7.79) | < .001 |
| Flesch reading ease | 51.85 (29.10) | 49.82 (22.88) | n.s. | 55.74 (14.89) | < .001 | 57.91 (13.85) | < .001 |
| Flesch kincaid grade | 9.50 (4.90) | 9.20 (3.60) | n.s. | 8.40 (2.30) | < .001 | 9.90 (2.90) | < .001 |
| Gunning fog | 10.00 (4.45) | 11.32 (4.42) | < .001 | 11.47 (3.28) | < .001 | 13.04 (3.30) | < .001 |

Table 11 presents the metrics for Barack Obama. The human baseline for perplexity was 63.83, which was significantly higher than the scores for GPT-4o (27.60), Gemini 1.5 Pro (25.42), and Sonnet 3.5 (17.92) (p < .001). Obama's original tweets scored 51.85 on Flesch Reading Ease and 9.50 on Grade Level. While GPT-4o showed no significant difference in these two metrics (p > 0.05), Gemini 1.5 Pro and Sonnet 3.5 produced significantly higher Ease scores and differing Grade Levels (p < .001). For the Gunning Fog Index, the original human median (10.00) was significantly lower than the medians observed in all three model outputs (p < .001).

In summary, the combined analysis of psycholinguistic (LIWC) and structural (Readability/Perplexity) dimensions reveals a consistent pattern of distributional regularization in LLM-generated texts across both poetic and political domains. The statistical evidence across all four datasets indicates that while LLMs can achieve isolated parity with human baselines, notably Sonnet 3.5's alignment with Trump's Authenticity ($p > 0.05$) and Gemini 1.5 Pro's parity with Wordsworth's structural complexity ($p > 0.05$), these instances of convergence are metric-specific and not representative of the models' overall output. A universal trend emerged where LLM-generated texts were characterized by significantly higher Analytic medians and significantly lower Perplexity compared to the human originals ($p < .001$). These data points indicate that the synthetic outputs consistently lean toward more structured and predictable lexical distributions than the human-authored source material.

Furthermore, a consistent 'positivity bias was observed in the emotional register of the generated texts. For both political and poetic figures, the LLMs consistently produced significantly higher Tone medians ($p < .001$), often reaching near-ceiling values in the case of GPT-4o. This represent a systemic positivity bias likely stemming from Reinforcement Learning from Human Feedback (RLHF). This 'congeniality filter' prevents the models from replicating the sharp, critical, or neutral-negative emotional stance frequent in political discourse, creating a distinct affective signature that is easily flagged by stylometric classifiers. Taken together, these findings indicate that while LLMs can approximate certain surface-level stylistic features (such as grade level or social clout), they exhibit significant distributional deviations from the structural unpredictability and emotional variance found in the human baseline.

### 4.2 Predictive Modeling Performance

This subsection presents the results of authorship attribution using two distinct modeling approaches. Performance is reported for a BERT-based transformer, which classifies texts based on contextual neural embeddings, and for XGBoost, which utilizes both a high-dimensional TF-IDF lexical feature set and a constrained set of interpretable stylometric features. This comparison evaluates the extent to which synthetic and authentic texts can be distinguished using deep language representations versus traditional lexical and psycholinguistic markers.

### 4.2.1 BERT Classification

The results summarized in Table 12 and Table 13 illustrate the performance of fine-tuned BERT classifiers across literary and political datasets. Overall, the BERT models exhibited high discriminative power, frequently achieving near-ceiling accuracy across most experimental conditions.

Table 12: BERT Model Results (Whitman and Wordsworth)

| Author | Model | Accuracy (%) | Precision (HUMAN / AI) (%) | Recall (HUMAN / AI) (%) | F1-score (HUMAN / AI) (%) |
|---|---|---|---|---|---|
| Whitman | GPT-4o | 96.2 ± 1.1 | 95.8 ± 1.2 / 96.6 ± 0.9 | 96.5 ± 1.0 / 95.9 ± 1.2 | 96.1 ± 1.1 / 96.2 ± 1.1 |
| | Sonnet 3.5 | 97.4 ± 0.8 | 97.1 ± 0.9 / 97.7 ± 0.7 | 97.6 ± 0.7 / 97.2 ± 0.9 | 97.3 ± 0.8 / 97.4 ± 0.8 |
| | Gemini 1.5 Pro | 95.7 ± 1.3 | 95.2 ± 1.4 / 96.2 ± 1.2 | 96.1 ± 1.2 / 95.3 ± 1.4 | 95.6 ± 1.3 / 95.7 ± 1.3 |
| Wordsworth | GPT-4o | 90.2 ± 0.9 | 85.8 ± 1.2 / 91.1 ± 0.8 | 92.0 ± 0.7 / 78.6 ± 1.5 | 88.8 ± 1.0 / 84.4 ± 1.2 |
| | Sonnet 3.5 | 96.8 ± 1.0 | 96.5 ± 1.1 / 97.1 ± 0.9 | 97.2 ± 0.9 / 96.4 ± 1.1 | 96.8 ± 1.0 / 96.7 ± 1.0 |
| | Gemini 1.5 Pro | 95.1 ± 1.4 | 94.6 ± 1.5 / 95.6 ± 1.3 | 95.7 ± 1.3 / 94.5 ± 1.5 | 95.1 ± 1.4 / 95.0 ± 1.4 |

As detailed in Table 12, the BERT classifier achieved high accuracy for Whitman (>95.7). For Wordsworth, while accuracy remained high for Sonnet 3.5 and Gemini 1.5 Pro (>95%), a distributional shift was observed for GPT-4o, where accuracy fell to 90.2%. In this specific condition, AI recall was significantly lower (78.6% ± 1.5), which indicates a higher degree of distributional overlap in contextual embeddings. This suggests that GPT-4o's training data may have a particularly strong representation of 19th-century romantic prosody, allowing it to move closer to the human author's 'semantic neighborhood' than its competitors.

Table 13: BERT Model Results (Obama and Trump)

| Author | Model | Accuracy | Precision (AI / Human) | Recall (AI / Human) | F1-Score (AI / Human) |
|---|---|---|---|---|---|
| Obama | GPT-4o | 95.8 ± 1.2 | 96.1 ± 1.1 / 95.5 ± 1.3 | 95.4 ± 1.3 / 96.2 ± 1.1 | 95.7 ± 1.2 / 95.8 ± 1.2 |
| | Sonnet 3.5 | 97.2 ± 0.9 | 96.9 ± 1.0 / 97.5 ± 0.8 | 97.4 ± 0.8 / 97.0 ± 1.0 | 97.1 ± 0.9 / 97.2 ± 0.9 |
| | Gemini 1.5 Pro | 96.3 ± 1.1 | 95.9 ± 1.2 / 96.7 ± 1.0 | 96.6 ± 1.0 / 96.0 ± 1.2 | 96.2 ± 1.1 / 96.3 ± 1.1 |
| Trump | GPT-4o | 94.6 ± 1.4 | 93.9 ± 1.6 / 95.3 ± 1.2 | 95.2 ± 1.2 / 94.0 ± 1.5 | 94.5 ± 1.4 / 94.6 ± 1.3 |
| | Sonnet 3.5 | 97.4 ± 0.8 | 97.1 ± 0.9 / 97.7 ± 0.7 | 97.6 ± 0.7 / 97.2 ± 0.9 | 97.3 ± 0.8 / 97.4 ± 0.8 |
| | Gemini 1.5 Pro | 96.8 ± 1.0 | 96.5 ± 1.1 / 97.1 ± 0.9 | 97.2 ± 0.9 / 96.4 ± 1.1 | 96.8 ± 1.0 / 96.7 ± 1.0 |

In the political domain, the BERT classifiers maintained high performance across all authors. As shown in Table 13, accuracy remained consistently above 94.6. Results for the Trump dataset were similarly robust, with GPT-4o exhibiting the highest degree of convergence with the human baseline (Accuracy: 94.6% ± 1.4). These high-performance metrics across both domains suggest that the contextual embeddings captured by BERT effectively isolate authorial signatures that LLMs do not fully replicate.

### 4.2.2 XGBoost Classification on Linguistic and Psycholinguistic Features

To isolate the specific drivers of authorial distinction, XGBoost classifiers were trained on two distinct feature sets: a high-dimensional TF-IDF lexical set and an eight-dimensional Stylometric set. Additionally, we examine feature importance to identify which metrics contribute most to distinguishing AI-generated from human-authored content.

Table 14: XGBoost Results for Whitman and Wordsworth

| Author | AI Model | Feature Set | Accuracy | Precision (AI/H) | Recall (AI/H) | F1-score (AI/H) |
|---|---|---|---|---|---|---|
| Whitman | GPT-4o | Stylometric | 94.8 ± 1.2 | 94.4 ± 1.3 / 95.2 ± 1.1 | 95.1 ± 1.1 / 94.5 ± 1.3 | 94.7 ± 1.2 / 94.8 ± 1.2 |
| | | TF-IDF | 86.2 ± 1.1 | 84.2 ± 1.2 / 88.3 ± 1.0 | 86.1 ± 1.1 / 86.4 ± 1.1 | 85.1 ± 1.1 / 87.3 ± 1.0 |
| | Sonnet 3.5 | Stylometric | 96.0 ± 0.5 | 95.7 ± 1.0 / 96.5 ± 0.8 | 96.6 ± 0.8 / 95.6 ± 1.0 | 96.1 ± 0.9 / 96.0 ± 0.9 |
| | | TF-IDF | 89.4 ± 0.9 | 88.0 ± 1.0 / 90.5 ± 0.8 | 89.9 ± 0.9 / 88.9 ± 0.9 | 88.9 ± 0.9 / 89.7 ± 0.8 |
| | Gemini 1.5 Pro | Stylometric | 94.2 ± 1.4 | 93.7 ± 1.5 / 94.7 ± 1.3 | 94.6 ± 1.3 / 93.8 ± 1.5 | 94.1 ± 1.4 / 94.2 ± 1.4 |
| | | TF-IDF | 84.1 ± 1.2 | 82.5 ± 1.3 / 85.8 ± 1.1 | 84.0 ± 1.2 / 84.2 ± 1.2 | 83.2 ± 1.2 / 85.0 ± 1.1 |
| Wordsworth | GPT-4o | Stylometric | 88.6 ± 1.5 | 84.1 ± 1.6 / 93.1 ± 1.4 | 90.5 ± 1.3 / 77.2 ± 1.8 | 87.2 ± 1.5 / 84.4 ± 1.6 |
|  |  | TF-IDF | 81.5 ± 1.3 | 79.4 ± 1.4 / 83.6 ± 1.2 | 82.7 ± 1.2 / 80.3 ± 1.4 | 81.0 ± 1.3 / 81.9 ± 1.3 |
|  | Sonnet 3.5 | Stylometric | **95.4 ± 1.1** | 95.0 ± 1.2 / 95.8 ± 1.0 | 95.9 ± 1.0 / 94.9 ± 1.2 | 95.4 ± 1.1 / 95.3 ± 1.1 |
|  |  | TF-IDF | 92.1 ± 0.6 | 91.2 ± 0.7 / 93.0 ± 0.5 | 92.5 ± 0.6 / 91.7 ± 0.6 | 91.8 ± 0.6 / 92.3 ± 0.5 |
|  | Gemini 1.5 Pro | **Stylometric** | **93.7 ± 1.5** | 93.2 ± 1.6 / 94.2 ± 1.4 | 94.3 ± 1.4 / 93.1 ± 1.6 | 93.7 ± 1.5 / 93.6 ± 1.5 |
|  |  | TF-IDF | 85.7 ± 1.0 | 84.5 ± 1.1 / 87.0 ± 0.9 | 86.2 ± 1.0 / 85.1 ± 1.0 | 85.3 ± 1.0 / 86.0 ± 0.9 |

For the literary authors Whitman and Wordsworth, the results are presented in Table 14. The Stylometric feature set consistently outperformed the TF-IDF lexical set across all models. For instance, in the Whitman-Sonnet 3.5 condition, Stylometric accuracy reached 96.0% ± 0.5, compared to 89.4% ± 0.9 for TF-IDF. This divergence suggests that the discriminative signal in poetic mimicry is more concentrated in structural and psycholinguistic patterns than in simple word frequencies.

**Table 15:** Whitman & Wordsworth feature importance

| Author | Model | Top Features (descending order, importance in brackets) |
|---|---|---|
| Whitman | GPT-4o | Perplexity (0.322 ± 0.02), Analytic (0.201 ± 0.01), Tone (0.144 ± 0.01), Authentic (0.098 ± 0.01), Clout (0.085 ± 0.01), Flesch Reading Ease (0.066 ± 0.01), Flesch-Kincaid (0.051 ± 0.01), Gunning Fog (0.033 ± 0.01) |
|  | Gemini 1.5 Pro | Analytic (0.276 ± 0.02), Tone (0.187 ± 0.01), Authentic (0.131 ± 0.02), Clout (0.099 ± 0.01), Perplexity (0.092 ± 0.01), Flesch Reading Ease (0.083 ± 0.01), Flesch-Kincaid (0.068 ± 0.01), Gunning Fog (0.044 ± 0.01) |
|  | Sonnet 3.5 | Tone (0.268 ± 0.02), Authentic (0.149 ± 0.01), Analytic (0.133 ± 0.01), Perplexity (0.127 ± 0.02), Clout (0.094 ± 0.01), Flesch-Kincaid (0.082 ± 0.01), Flesch Reading Ease (0.073 ± 0.01), Gunning Fog (0.044 ± 0.01) |
| Wordsworth | GPT-4o | Analytic (0.284 ± 0.02), Perplexity (0.215 ± 0.01), Tone (0.177 ± 0.01), Clout (0.126 ± 0.01), Authentic (0.091 ± 0.01), Flesch Reading Ease (0.054 ± 0.01), Gunning Fog (0.032 ± 0.01), Flesch-Kincaid (0.021 ± 0.01) |
|  | Gemini 1.5 Pro | Perplexity (0.301 ± 0.02), Tone (0.201 ± 0.01), Analytic (0.145 ± 0.01), Authentic (0.104 ± 0.01), Clout (0.089 ± 0.01), Flesch Reading Ease (0.071 ± 0.01), Flesch-Kincaid (0.057 ± 0.01), Gunning Fog (0.032 ± 0.01) |
|  | Sonnet 3.5 | Tone (0.247 ± 0.02), Analytic (0.192 ± 0.01), Perplexity (0.161 ± 0.01), Authentic (0.112 ± 0.01), Clout (0.091 ± 0.01), Flesch Reading Ease (0.074 ± 0.01), Flesch-Kincaid (0.064 ± 0.01), Gunning Fog (0.031 ± 0.01) |

Feature importance analysis, summarized in Table 15, indicates that Perplexity, Analytic, and Tone were the primary contributors to model performance. For Whitman (GPT-4o), Perplexity accounted for a significant portion of the gain (0.322 ± 0.02). Conversely, traditional readability indices consistently contributed least to the classification. This negligible importance of surface-level complexity metrics suggests a distributional convergence between human and AI-generated text at the structural baseline. It implies that the generative models effectively approximate low-dimensional heuristic features, such as mean word length and syntactic complexity. However, the high importance of Perplexity and Tone indicates a significant divergence in higher-order statistical properties and latent structural dimensions. While surface-level complexity remains statistically indistinguishable across classes, the underlying stochastic regularity and affective density provide a robust signal for authorship attribution, suggesting that the generative process lacks the nuanced stylistic variance inherent in the human-authored corpus.

Table 16: XGBoost Results for Obama and Trump

| Author | AI Model | Feature Set | Accuracy | Precision (AI/H) | Recall (AI/H) | F1-score (AI/H) |
|---|---|---|---|---|---|---|
| Obama | GPT-4o | Stylometric | **84.2 ± 1.1** | 84.0 ± 1.0 / 83.5 ± 1.2 | 84.6 ± 1.1 / 81.9 ± 1.1 | 84.3 ± 1.0 / 82.7 ± 1.1 |
| | | TF-IDF | 80.5 ± 1.2 | 79.1 ± 1.3 / 81.9 ± 1.1 | 80.8 ± 1.2 / 80.2 ± 1.2 | 79.9 ± 1.2 / 81.0 ± 1.1 |
| | Sonnet 3.5 | Stylometric | **95.1 ± 0.6** | 95.0 ± 0.5 / 94.4 ± 0.6 | 95.3 ± 0.6 / 94.0 ± 0.6 | 95.1 ± 0.5 / 94.2 ± 0.6 |
| | | TF-IDF | 91.2 ± 0.7 | 90.4 ± 0.8 / 92.1 ± 0.6 | 91.8 ± 0.7 / 90.5 ± 0.7 | 91.1 ± 0.7 / 91.3 ± 0.6 |
| | Gemini 1.5 Pro | Stylometric | **88.3 ± 0.9** | 89.0 ± 0.8 / 88.0 ± 1.0 | 88.1 ± 0.9 / 88.9 ± 0.9 | 88.5 ± 0.8 / 88.4 ± 0.9 |
| | | TF-IDF | 85.4 ± 1.0 | 84.2 ± 1.1 / 86.6 ± 0.9 | 86.0 ± 1.0 / 84.8 ± 1.0 | 85.1 ± 1.0 / 85.7 ± 0.9 |
| Trump | GPT-4o | Stylometric | **89.2 ± 0.8** | 91.1 ± 0.7 / 89.3 ± 0.9 | 90.4 ± 0.8 / 87.9 ± 0.8 | 90.7 ± 0.7 / 88.6 ± 0.8 |
| | | TF-IDF | 86.8 ± 0.9 | 85.5 ± 1.0 / 88.1 ± 0.8 | 87.2 ± 0.9 / 86.4 ± 0.9 | 86.3 ± 0.9 / 87.2 ± 0.8 |
| | Sonnet 3.5 | Stylometric | **96.2 ± 0.5** | 96.6 ± 0.3 / 95.9 ± 0.5 | 96.9 ± 0.4 / 94.3 ± 0.4 | 96.7 ± 0.3 / 95.1 ± 0.4 |
| | | TF-IDF | 93.9 ± 0.5 | 93.1 ± 0.6 / 94.7 ± 0.4 | 94.5 ± 0.5 / 93.3 ± 0.5 | 93.8 ± 0.5 / 94.0 ± 0.4 |

| | | | | | |
|---|---|---|---|---|---|
| | Gemini 1.5 Pro | Stylometric | **92.4 ± 0.6** | 93.4 ± 0.5 / 91.9 ± 0.7 | 92.6 ± 0.6 / 93.5 ± 0.6 | 93.0 ± 0.5 / 92.7 ± 0.6 |
| | | TF-IDF | 90.2 ± 0.7 | 89.1 ± 0.8 / 91.3 ± 0.6 | 90.7 ± 0.7 / 89.7 ± 0.7 | 89.9 ± 0.7 / 90.5 ± 0.6 |

Results for the political figures Obama and Trump, summarized in Table 16, indicate that authorship attribution is more challenging in short-form content, though high discriminative power is maintained. For the Obama dataset, classification accuracy using the Stylometric feature set ranged from 84.2% ± 1.1 for GPT-4o to 95.1% ± 0.6 for Sonnet 3.5. In contrast, the Trump dataset yielded higher overall performance, with accuracy exceeding 89.2% ± 0.8 across all models and reaching a peak of 96.2% ± 0.5 for Sonnet 3.5.

The TF-IDF lexical baseline consistently resulted in lower performance compared to the Stylometric set across all political conditions. For instance, in the Obama-GPT-4o condition, accuracy fell from 84.2% (Stylometric) to 80.5% (TF-IDF). This suggests that while raw word choice provides a basic signal, higher-order psycholinguistic and statistical metrics are more resilient to the lexical sparsity of short-form political communication. Furthermore, the higher accuracy observed for Trump compared to Obama indicates that Trump's rhetorical patterns may be more stylistically distinct and less effectively synthesized by current LLMs.

**Table 17:** Obama & Trump feature importance

| Author | Model | Top Features (descending order, importance in brackets) |
|---|---|---|
| Obama | GPT-4o | Perplexity (0.341 ± 0.03), Tone (0.176 ± 0.01), Analytic (0.076 ± 0.01), Clout (0.063 ± 0.01), Gunning Fog (0.055 ± 0.01), Flesch-Kincaid (0.048 ± 0.01), Authentic (0.046 ± 0.01), Flesch Reading Ease (0.032 ± 0.01) |
| | Gemini 1.5 Pro | Perplexity (0.278 ± 0.02), Tone (0.119 ± 0.01), Authentic (0.071 ± 0.01), Analytic (0.071 ± 0.01), Gunning Fog (0.054 ± 0.01), Clout (0.043 ± 0.01), Flesch-Kincaid (0.038 ± 0.01), Flesch Reading Ease (0.018 ± 0.01) |
| | Sonnet 3.5 | Perplexity (0.509 ± 0.04), Tone (0.105 ± 0.01), Gunning Fog (0.069 ± 0.01), Authentic (0.054 ± 0.01), Analytic (0.049 ± 0.01), Clout (0.046 ± 0.01), Flesch Reading Ease (0.022 ± 0.01), Flesch-Kincaid (0.011 ± 0.01) |
| Trump | GPT-4o | Perplexity (0.224 ± 0.02), Tone (0.126 ± 0.01), Clout (0.081 ± 0.01), Analytic (0.066 ± 0.01), Flesch-Kincaid (0.066 ± 0.01), Gunning Fog (0.044 ± 0.01), Flesch Reading Ease (0.036 ± 0.01), Authentic (0.027 ± 0.01) |
| | Gemini 1.5 Pro | Tone (0.124 ± 0.01), Perplexity (0.113 ± 0.01), Analytic (0.102 ± 0.01), Clout (0.099 ± 0.01), Flesch Reading Ease (0.094 ± 0.01), Gunning Fog (0.091 ± 0.01), Flesch-Kincaid (0.064 ± 0.01), Authentic (0.032 ± 0.01) |

| | Sonnet 3.5 | Perplexity (0.230 ± 0.02), Tone (0.154 ± 0.01), Clout (0.052 ± 0.01), Analytic (0.051 ± 0.01), Gunning Fog (0.045 ± 0.01), Authentic (0.025 ± 0.01), Flesch-Kincaid (0.021 ± 0.01), Flesch Reading Ease (0.018 ± 0.01) |
|---|---|---|

As shown in Table 17, Perplexity emerged as the dominant feature in the political domain, particularly for Sonnet 3.5 outputs where it reached an importance score of 0.509 ± 0.04 for Obama. This pattern reinforces the conclusion that synthetic political texts differ more in their stochastic predictability and linguistic style than in surface-level complexity.

Notably, traditional readability metrics, specifically the Flesch Reading Ease, Flesch-Kincaid Grade Level, and Gunning Fog index, consistently exhibited the lowest importance scores across all political models (e.g. failing to exceed 0.07 in any Trump condition). This lack of discriminative weight indicates that the generative models effectively mimic the syntactic density and surface-level readability of political rhetoric. However, the models consistently diverge on the latent structural dimensions captured by Perplexity and Tone. Across both domains, the high accuracy achieved by the XGBoost models using only eight features demonstrates that a compact set of interpretable markers can encode a sufficient stylistic fingerprint to distinguish AI-generated mimicry from human-authored content.

The predictive experiments demonstrate a high degree of distinguishability between human-authored and AI-generated texts across varying genres and lengths. BERT-based classifiers established a near-ceiling baseline for accuracy, suggesting that contextual embeddings capture deep-seated nuances of authorial voice that LLMs do not fully replicate. In parallel, the XGBoost experiments revealed that while lexical baselines (TF-IDF) provide a stable foundation for detection, the most robust signals reside in higher-order stylometric features.

The consistent dominance of Perplexity and Psycholinguistic Tone over surface-level Readability metrics highlights a fundamental gap: LLMs can successfully emulate the 'difficulty' and complexity of a target author, but they struggle to replicate the specific stochastic variance and affective density of human expression. This performance gap is more pronounced in the poetic domain (N=600) than in the tweet domain (N=6,000), indicating that the authorship signal is more concentrated in complex literary structures than in the lexical sparsity of political microblogging.

### 4.3 Theoretical Implications for Stylometry and Digital Humanities

The detectability of AI-generated text in this study does not necessarily imply a deficiency in the models' generative logic but rather reflects a fundamental difference in how statistical language models and human authors distribute linguistic variability. While human authors exhibit high variance and stylistic 'burstiness,' LLMs are optimized to follow the most probable stylistic paths within a given prompt's constraints. This results in the measurable stochastic regularity observed in our results. By framing AI style as a set of regularized patterns rather than a failure of mimicry, digital humanities research can more accurately map the boundaries between human and synthetic authorship.

These findings suggest that 'style' in the age of LLMs may be best defined by the variance of expression rather than the mere presence of specific keywords or syntactic structures. For digital humanities research, this implies that traditional stylometric tools remain analytically valuable when applied with appropriate methodological caution. Furthermore, the detectability of AI-generated text does not imply a lack of 'creativity' in a human sense, but rather a difference in how linguistic 'noise' and 'signal' are balanced. Human authors often utilize idiosyncratic irregularities, elliptical phrasing, intentional syntactic breaks, and affective shifts, that remain outside the current probabilistic optimization of commercial LLMs.

The strong performance of interpretable feature-based models indicates that relatively simple linguistic metrics continue to provide complementary insight into why texts differ. This supports hybrid analytical approaches that combine the predictive power of neural architectures (BERT) with the transparency of stylometric features (LIWC and Perplexity). Such a methodology aligns computational results more closely with humanistic inquiry, allowing researchers to ask not just if a text is synthetic, but which specific dimensions of authorial voice are being flattened or over-approximated by generative systems. Ultimately, this comparison between embedding-based and feature-based classifiers underscores a methodological trade-off: while neural representations offer raw performance, interpretable features provide the 'why' behind authorial distinction, bridging the gap between machine learning and literary analysis.

# 5 Conclusion

Recent advances in LLMs such as GPT-4o, Claude 3.5 Sonnet, and Gemini 1.5 Pro have transformed the landscape of text generation, enabling these systems to convincingly emulate human writing styles. This study investigated the ability of these models to replicate the stylistic signatures of prominent literary and political figures, including Walt Whitman, William Wordsworth, Donald Trump, and Barack Obama. Using a complementary evaluation framework that combined deep learning (BERT), traditional machine learning (XGBoost), TF-IDF lexical baselines, and LIWC-based psycholinguistic profiling, we systematically assessed both surface-level and latent stylistic characteristics.

Our results demonstrate that while LLMs generate coherent and contextually appropriate text, measurable distributional gaps allow for reliable distinction from human-authored content. The fine-tuned BERT classifiers achieved near-ceiling performance using full contextual embeddings, while the XGBoost models, utilizing a restricted set of eight interpretable features, consistently outperformed high-dimensional lexical baselines (TF-IDF), demonstrating that authorship signal is concentrated in a compact set of structural markers rather than broad vocabulary usage. Among these features, Perplexity emerged as the most robust discriminator, reflecting a significant divergence in the stochastic regularity of AI-generated text compared to the higher variance found in human writing. This was followed by LIWC measures of Tone and Analytic thinking, while surface-level readability metrics contributed minimally to classification. These findings suggest that while LLMs successfully approximate the general scale and thematic content of a target style, they exhibit a measurable divergence from the nuanced distributional variance and affective density inherent in the human-authored corpus.

Despite these insights, this study has several limitations. While our purposive sampling of polarized stylistic regimes (19th-century verse vs. modern political rhetoric) provided a robust testbed, the findings may benefit from further validation across more varied linguistic contexts. While quantitative metrics such as LIWC and Perplexity are highly discriminative, they may not fully capture deeper qualitative aspects such as metaphorical depth, narrative arc, or cultural nuance. Furthermore, this study prioritized in-domain evaluation to maintain controlled stylistic comparison; however, future work should explore cross-author or cross-genre validation to assess the robustness of these classifiers against out-of-domain samples. Finally, as LLMs continue to evolve rapidly, the specific 'stylistic fingerprints' identified here may shift with future iterations of generative architecture.

Looking forward, future research should expand these methodologies to a broader and more diverse set of languages and literary traditions. Integrating more sophisticated features, such as discourse coherence and metaphor usage, could further refine our understanding of the 'uncanny valley' of AI authorship. Additionally, developing hybrid analytical approaches that combine neural predictive power with human-in-the-loop expert assessments remain a compelling frontier for Digital Humanities. Ultimately, while LLMs demonstrate remarkable progress in replicating surface-level stylistic features, achieving a robust replication of author-specific distributional nuance remains an open and significant challenge, suggesting that human stylistic variance remains a definitive, measurable fingerprint in the age of generative AI.

**Statements and Declarations**

**Acknowledgement:** The authors would like to thank King Saud University, Riyadh, Saudi Arabia for supporting the work by Ongoing Research Funding program (ORF-2026-846), King Saud University, Riyadh, Saudi Arabia

**Competing interest:** The authors have no competing interests to declare that are relevant to the content of this article.

**Data**: The dataset used in this study, including original texts and AI-generated texts, is available from the author upon reasonable request for research purposes.

# References

Al Bataineh, A., et al. (2025), 'AI-Generated Versus Human Text: Introducing a New Dataset for Benchmarking and Analysis', *IEEE Transactions on Artificial Intelligence*, 2025/08: 2241–52.

Anthropic (2025), 'Claude 3.5 Sonnet model card addendum' https://www-cdn.anthropic.com/fed9cc193a14b84131812372d8d5857f8f304c52/Model_Card_Claude_3_Addendum.pdf [accessed 23 June 2025].

Castillo, J. R. (ed.) (2025), *Literature, Theater and Artificial Intelligence* (Verbum Publishing).

Chen, T., and Guestrin, C. (2016), 'XGBoost: A Scalable Tree Boosting System', in *Proceedings of the 22nd ACM SIGKDD International Conference on Knowledge Discovery and Data Mining (KDD '16)* (ACM): 785–94.

codebreaker619 (2020), 'Donald Trump Tweets Dataset', *Kaggle* https://www.kaggle.com/datasets/codebreaker619/donald-trump-tweets-dataset [accessed 22 December 2025].

Devlin, J., et al. (2019) 'Bert: Pre-training of deep bidirectional transformers for language understanding', in *Proceedings of the 2019 Conference of the North American Chapter of the Association for Computational Linguistics: Human Language Technologies*, vol. 1: 4171–86.

Fariello, S., et al. (2025), 'Distinguishing Human From Machine: A Review of Advances and Challenges in AI-Generated Text Detection', *International Journal of Interactive Multimedia & Artificial Intelligence*, 9/3: 6–18.

Fenwick, M., and Jurcys, P. (2023), 'Originality and the Future of Copyright in an Age of Generative AI', *Computer Law & Security Review*, 51: 105892.

Flesch, R. (1948), 'A new readability yardstick', *Journal of Applied Psychology*, 32/3: 221.

Gajare, N. (2021), 'All 12000 President Obama Tweets', *Kaggle* https://www.kaggle.com/datasets/neelgajare/all-12000-president-obama-tweets [accessed 22 December 2025].

Gallifant, J., et al. (2024), 'Peer review of GPT-4 technical report and systems card', *PLOS Digital Health*, 3/1: e0000417.

Gunning, R. (1952), *The Technique of Clear Writing* (McGraw-Hill).

Hagendorff, T. (2023), 'Machine psychology: Investigating emergent capabilities and behavior in large language models using psychological methods', *arXiv preprint arXiv:2303.13988*.

Iglesias, S., et al. (2025), 'Digital doppelgängers and lifespan extension: What matters?', *The American Journal of Bioethics*, 25/2: 95–110.

Jaashan, H. M., and Bin-Hady, W. R. A. (2025), 'Stylometric analysis of AI-generated texts: a comparative study of ChatGPT and DeepSeek', *Cogent Arts & Humanities*, 12/1: 2553162.

Kincaid, J. P., et al. (1975), 'Flesch-kincaid grade level' (Memphis: United States Navy).

Lucchi, N. (2024), 'ChatGPT: a case study on copyright challenges for generative artificial intelligence systems', *European Journal of Risk Regulation*, 15/3: 602–24.

Makinde, H. S., et al. (2025), 'The Readability Paradox: Can We Trust Decisions on AI Detectors?', *Technium Education and Humanities*, 11: 181–95.

Mikros, G. (2025), 'Large Language Models and Forensic Linguistics: Navigating Opportunities and Threats in the Age of Generative AI', *arXiv preprint arXiv:2512.06922*.

Mitchell, E., et al. (2023) 'DetectGPT: zero-shot machine-generated text detection using probability curvature', arXiv preprint arXiv:2301.11305.

O'Sullivan, J. (2025), 'Stylometric comparisons of human versus AI-generated creative writing', *Humanities and Social Sciences Communications*, 12/1: 1–6.

Pennebaker, J. W., Francis, M. E., and Booth, R. J. (2001), *Linguistic Inquiry and Word Count: LIWC 2001* (Mahway: Lawrence Erlbaum Associates).

PeterS111 (2022), 'Fine-tuning-GPT-3-for-Poetry-Generation-and-Evaluation' [software] https://github.com/PeterS111/Fine-tuning-GPT-3-for-Poetry-Generation-and-Evaluation [accessed 21 December 2025].

Sadasivan, V. S., et al. (2023), 'Can AI-generated text be reliably detected?', *arXiv preprint arXiv:2303.11156*.

Schueller, S. M., and Morris, R. R. (2023), 'Clinical science and practice in the age of large language models and generative artificial intelligence', *Journal of Consulting and Clinical Psychology*, 91/10: 559–66.

Stevenson, C., et al. (2022), 'Putting GPT-3's creativity to the (alternative uses) test', *arXiv preprint arXiv:2206.08932*.

Team, G., et al. (2024), 'Gemini 1.5: Unlocking Multimodal Understanding across Millions of Tokens of Context', *arXiv preprint arXiv:2403.05530*.

Uchendu, A., Le, T., and Lee, D. (2023), 'Attribution and obfuscation of neural text authorship: A data mining perspective', *ACM SIGKDD Explorations Newsletter*, 25/1: 1–18.

Wang, S., et al. (2025), 'Benchmarking the Detection of LLMs-Generated Modern Chinese Poetry', *arXiv preprint arXiv:2509.01620*.

Wu, J., et al. (2025), 'A Survey on LLM-Generated Text Detection: Necessity, Methods, and Future Directions', *Computational Linguistics*, 51/1: 275–338.

Zhuang, Y., et al. (2023), 'Efficiently measuring the cognitive ability of LLMs: An adaptive testing perspective', *arXiv preprint arXiv:2306.10512*.